\def\year{2021}\relax
\documentclass[letterpaper]{article} 
\usepackage{aaai21}  
\usepackage{times}  
\usepackage{helvet} 
\usepackage{courier}  
\usepackage[hyphens]{url}  
\usepackage{graphicx} 
\usepackage{natbib}  
\usepackage{caption} 
\usepackage{amsmath,amssymb, bbm}
\usepackage{algorithm}
\usepackage{multirow}
\usepackage{siunitx}
\usepackage[noend]{algpseudocode}
\usepackage{appendix}
\usepackage{booktabs}
\usepackage{cleveref}

\usepackage[switch]{lineno}

\crefname{section}{section}{sections}
\newcommand{\Sspace}{\mathcal{S}}
\newcommand{\Aspace}{\mathcal{A}}
\newcommand{\Ospace}{\mathcal{O}}

\title{Bayesian Optimized Monte Carlo Planning}
\author{John Mern,\textsuperscript{\rm 1} 
        Anil Yildiz,\textsuperscript{\rm 1} 
        Zachary Sunberg,\textsuperscript{\rm 2} 
        Tapan Mukerji,\textsuperscript{\rm 3} 
        Mykel J. Kochenderfer\textsuperscript{\rm 1} \\}
\affiliations {
    \textsuperscript{\rm 1}Stanford University, Department of Aeronautics and Astronautics, 496 Lomita Mall, Stanford, CA 94305 \\
    \textsuperscript{\rm 2}University of Colorado Boulder, Department of Aerospace Engineering Sciences, 3775 Discovery Drive, Boulder, CO 80303 \\
    \textsuperscript{\rm 3}Stanford University, Department of Energy Resources Engineering, 367 Panama Street, Stanford, CA 94305 \\
    \{jmern91, yildiz, mukerji, mykel\}@stanford.edu \\
    zachary.sunberg@colorado.edu
}

\begin{document}
\maketitle

\begin{abstract}
Online solvers for partially observable Markov decision processes have difficulty scaling to problems with large action spaces. 
Monte Carlo tree search with progressive widening attempts to improve scaling by sampling from the action space to construct a policy search tree.
The performance of progressive widening search is dependent upon the action sampling policy, often requiring problem-specific samplers. 
In this work, we present a general method for efficient action sampling based on Bayesian optimization.
The proposed method uses a Gaussian process to model a belief over the action-value function and selects the action that will maximize the expected improvement in the optimal action value.
We implement the proposed approach in a new online tree search algorithm called Bayesian Optimized Monte Carlo Planning (BOMCP). 
Several experiments show that BOMCP is better able to scale to large action space POMDPs than existing state-of-the-art tree search solvers. 

\end{abstract}

\section{Introduction}
The Partially Observable Markov Decision Process (POMDP) is a mathematical model of sequential decision making problems under state uncertainty~\cite{littman1995}.
In a POMDP, an agent takes actions while receiving noisy observations of the world state in order to accumulate reward.
A policy that solves a POMDP maps histories of observations, represented as a belief over the state, to actions that maximize the expected sum of discounted rewards.

Solving POMDPs exactly is generally intractable and has been shown to be PSPACE-\textit{complete} over finite horizons~\cite{papadimitriou1987}.
A policy for a POMDP with $n$ distinct states requires a function over an $(n - 1)$-dimensional continuous belief space~\cite{Silver2010}.
Computing exact solutions over this space requires evaluating a number of potential trajectories that is exponential in the time horizon~\cite{pineau2006}.
These phenomena are referred to as the \emph{curse of dimensionality} and the \emph{curse of history}, respectively. 

Because of these challenges, sample-based methods are often used to solve POMDPs approximately~\cite{smith2004,kurniawati2008}. 
Methods based on Monte Carlo Tree Search (MCTS) have been particularly effective in recent years as online-solvers.
MCTS methods work by incrementally building a search tree over trajectories reachable from the agent's current belief and using the resulting tree to estimate the value of the available actions. 
While tree-search methods typically handle POMDPs with large state spaces well, they often perform poorly on problems with large action spaces~\cite{browne2012survey}.

Large action and observation spaces tend to lead to excessive branching in approximate tree-search solvers such as POMCP~\cite{Silver2010}, resulting in shallow trees with poor value estimates.
Progressive widening limits tree-width by sampling a subset of the available actions for addition to the search tree~\cite{couetoux2011}.
Random action sampling, however, may often fail to select good actions for evaluation, requiring expert-biased sampling for acceptable performance~\cite{browne2012}.

This work frames progressive action selection as an optimization problem to be solved within MCTS.
The proposed method models the problem at each branching step using a Bayesian optimization framework~\cite{mockus2012}.
Under this framework, the optimization objective is to select the action with the highest expected improvement in the action-value returned by the subsequent MCTS search.
To calculate expected improvement, a distribution over the unknown action-value function is modeled as a Gaussian Process~\cite{rasmussen2005}.
The procedure uses knowledge gained during tree search to guide action selection by conditioning the Gaussian Process on the values of nodes already in the tree.

We implemented the proposed method in a new MCTS planner called Bayesian Optimized Monte Carlo Planning (BOMCP). 
The BOMCP algorithm builds on the state-of-the-art solver, POMCPOW~\cite{sunberg2018}, to better scale to large action spaces. 
To efficiently scale the Bayesian optimization procedure, we introduce a $k$-nearest neighbor method of Gaussian Process inference to model the value function distribution in the Bayesian Optimizing procedure. 
Experiments show that this new algorithm outperforms POMCPOW, on several problems with large action-spaces.

\section{Related Work}

There has been significant work in online planning for POMDPS with Monte Carlo Tree Search~\cite{browne2012survey}. 
One of the first successful planners, POMCP~\cite{Silver2010} adapted UCT exploration~\cite{kocsis2006} by sampling states from an unweighted particle set and searching over action-observation trajectories. 
POMCP cannot scale well to problems with large action or observation spaces due to its branching strategy that leads to high tree width. 

DESPOT~\cite{somani2013} builds on POMCP by using a deterministic generative model to sample scenarios, resulting in a sparse search tree.
Due to the reduced observation branching, the number of nodes in the tree is significantly decreased and value estimates are improved. 
However, DESPOT can also struggle with large action-space problems without an expert action branching strategy. 

Double Progressive Widening (DPW)~\cite{couetoux2011} was introduced to scale MDP planners to large discrete and continuous spaces.
Progressive widening dynamically limits the number of child nodes that may be added to a parent node based on the number of times the parent has been visited in search. 
DPW, however, has been shown to be sensitive to the order nodes are selected for addition~\cite{browne2012} and has limited effect on scaling to \textit{very} large action spaces on its own.

The POMCPOW and PFT-DPW algorithms~\cite{sunberg2018} introduce double progressive widening for POMDPs with large action and observation spaces. 
The algorithms maintain weighted particle beliefs at internal nodes to prevent the degeneration of multiple beliefs to a single particle.
However, due to the random sampling used during action branching, POMCPOW can still struggle on large scale problems. In particular, \citeauthor{sunberg2018} only apply POMCPOW to an action space with a single continuous dimension.

Work has been done to develop intelligent action selection methods for progressive widening for a variety of specialized cases. 
One method suggests using MCTS for a global search of coarsely discretized actions, while applying a finer local search to more promising actions, using value gradients~\cite{lee2020monte}. 
The value gradient calculation was found to be computationally expensive for many problems.

Another method uses Voronoi partitioning to dynamically cluster large action spaces and search by sampling from the Voronoi cells~\cite{kim2020monte}. 
However, cells are created at random locations and diameters and do not use knowledge gained during search. 

The Kernel Regression UCT algorithm (KR-UCT)~\cite{yee2016} introduces information sharing between action nodes of an MCTS tree through smooth kernel regression on value estimates.
KR-UCT, however, is restricted to POMDPs with uncertainty in action execution.

The Continuous Belief Tree Search algorithm (CBTS)~\cite{morere2016bayesian} selects actions using a Bayesian optimization process.
CBTS, however, does not use a progressive widening approach, and instead adds actions up to a fixed limit within a fully observable MCTS solver.
CBTS selects actions that maximize the return upper confidence bound, not the expected return improvement, and uses a separate Gaussian Process for each belief node. 
This restricts how much the action selection process can learn from value estimates at other belief nodes.

Finally, GPS-ABT~\cite{seiler2015online} uses generalized pattern search to find local optima in POMDPs with discrete observation spaces.

\section{Background}
This section reviews several topics that are foundational to the remainder of the paper. 
The main discussion assumes familiarity with POMDPs, Monte Carlo Tree Search, and Gaussian Processes, which are all briefly reviewed here. 
\subsection{POMDPs}
Partially Observable Markov Decision Processes (POMDPs) are compact representations of sequential decision problems with uncertainty~\cite{kochenderfer2015}. 
The problem state evolves according to probabilistic dynamics that are conditioned only on the current state and action. 
In a POMDP, the agent is assumed to only have access to noisy observations of the state.
Using these observations, the agent's task is to plan a sequence of actions that maximizes the total accumulated reward over the problem horizon. 

Formally, a POMDP is defined by a tuple $ (\Sspace, \Aspace, \Ospace, Z, T, R, \gamma) $, where $\Sspace$ is the state space, $\Aspace$ is the action space, and $\Ospace$ is the observation space. 
The transition model $\mathrm{T}(s' \mid s,a)$ gives the probability of transitioning from state $s$ to state $s'$ after taking action $a$.
The reward function $\mathrm{R}(s,a)$ specifies the immediate reward obtained after taking action $a$ at state $s$.
The observation model $\mathrm{Z}(o \mid s,a,s')$ specifies the probability of receiving observation $o$ in state $s'$ given that action $a$ had been taken at the previous state $s$. 
The $\gamma \in \left[0,1\right]$ is a time discount factor.

When solving a POMDP, it is common to maintain a distribution over the state, called the belief $b$.
The belief is updated each time the agent takes an action $a$ and receives an observation $o$, typically with a Bayesian update filter.


In a POMDP, there exists an optimal policy $\pi^*(b)$, which specifies the best action to take in any belief state $b$.
For a given policy, the action-value function $Q(b,a)$ gives the expected utility of taking action $a$ at belief state $b$ and then continuing to act according to the policy.
An optimal policy selects an action that maximizes $Q(b,a)$, from belief $b$.

\subsection{Monte Carlo Tree Search}

Most online POMDP solvers employ sample based tree search methods typically referred to as Monte-Carlo tree search. 
Partially observable MCTS methods construct search trees as alternating layers of actions taken and observations received, with a root at the current belief. 
The tree is then used to estimate the value of actions the agent can take from its current state.

A typical MCTS process proceeds by sampling a state from the agent's current belief. 
The state is used in a generative model to simulate potential trajectories. 
The simulation proceeds down the tree, selecting the best action from the existing branches until a leaf node is reached. 
It then adds a child node to the leaf and  performs a rollout using a baseline policy to estimate the value.
The value is propagated back through the parent nodes to the root to update their value estimates. 
Once a specified number of simulations have been run, the search returns the root action with the highest estimated value. 
Deeper trees tend to result in better estimates of the true optimal value. 

Many modern MCTS approaches employ double progressive widening (DPW) to dynamically limit tree width for better value estimates. 
In DPW, when a node $x$ is visited, a child node is added if 
\begin{equation}
    |\mathrm{Ch}(x)| \leq kN(x)^\alpha
\end{equation}
where $|\mathrm{Ch}(x)|$ is the number of children of node $x$, $N(x)$ is the number of visits to node $x$, and $k$ and $\alpha$ are hyperparameters. 
Separate $k$ and $\alpha$ pairs are defined for observation and action progressive widening.

A common approach to DPW is to select actions randomly for addition to the tree during expansion. 
This approach is typically only effective for problems with small, low-dimensional action spaces.
For problems with large action spaces, it is unlikely that the optimal action will be sampled and added to the tree. 
To overcome this, expertly biased sampling strategies are often required for each problem to be solved. 

\subsection{Gaussian Processes}
A Gaussian Process (GP) is a stochastic process which generates observations $y$ given inputs $x$ such that any subset of observations $\mathbf{y}$ is distributed according to a multivariate Gaussian distribution. 
Representing a set of observations from a GP as the combination of two subsets $\mathbf{y}$ and $\mathbf{y}^*$, the GP assumes
\begin{equation}
    \begin{bmatrix}
    \mathbf{y} \\
    \mathbf{y}^*
    \end{bmatrix} 
    \sim \mathcal{N}\Big(
    \begin{bmatrix}
        m(\mathbf{x}) \\
        m(\mathbf{x}^*) 
    \end{bmatrix}
    ,
    \begin{bmatrix}
    \mathbf{\Sigma}_{xx} & \mathbf{\Sigma}_{xx^*} \\
    \mathbf{\Sigma}_{x^*x} & \mathbf{\Sigma}_{x^*x^*} \\
    \end{bmatrix}
    + \mathbb{I}\sigma^2
    \Big)
\end{equation}
where $m$ is a scalar function of the input $x$ representing the prior mean. 
In practice, a constant mean is often used. 
The $\mathbf{\Sigma}$ matrices are the marginal and cross covariance matrices that compose the total covariance matrix. 
The covariance matrices are generated using a kernel function $K$ such that $\mathbf{\Sigma}_{ij} = K(x_i, x_j)$.
Observation noise is added as the $\sigma^2$ term. 

A typical use of a Gaussian Process is to model the conditional distribution of unobserved data given some previously observed data. 
Taking $\mathbf{y}$ and $\mathbf{x}$ to be our previously observed data and $\mathbf{y}^*$ and $\mathbf{x}^*$ to be the new query points, the conditional distribution takes the form
\begin{align}~\label{eq:GP Posterior}
    &P(\mathbf{y}^* \mid \mathbf{y}, \mathbf{x}, \mathbf{x}^* ) = \mathcal{N}(\mathbf{y} \mid \mathbf{\mu}^*, \mathbf{\Sigma}^*), \\
    &\mathbf{\mu}^* = m(\mathbf{x}^*) + \mathbf{\Sigma}_{x^*x}(\mathbf{\Sigma}_{xx} + \mathbb{I}\sigma^2)^{-1}(\mathbf{y} - m(\mathbf{x}))~\label{cond mu} \\
    &\mathbf{\Sigma}^* = \mathbf{\Sigma}_{x^*x^*} - \mathbf{\Sigma}_{x^*x}(\mathbf{\Sigma}_{xx} + \mathbb{I}\sigma^2)^{-1}\mathbf{\Sigma}_{x^*x}'~\label{cond sigma}
\end{align}

The above distributions may be used for calculating likelihoods or sampling new data at previously unobserved points.
\section{BOMCP}~\label{sec:BOMCP}
Action progressive widening approaches in MCTS are known to require effective action sampling approaches in order to achieve acceptable performance on tasks with large action spaces.
In this work, we introduce a general method for action sampling that uses the information learned during earlier steps of the tree-search process to select subsequent actions for tree expansion.
We present it in a new algorithm called Bayesian Optimized Monte Carlo Planning (BOMCP). 

The main innovation of BOMCP is the use of Bayesian optimization to select actions for progressive widening during Monte Carlo Tree Search. 
Bayesian optimization is a black-box optimization approach that uses a probabilistic model to optimize over an unknown target function.
In this case, the unknown optimization target is the action-value function $Q(b,a)$. 
Each time a new action is selected during tree-expansion, BOMCP selects the optimal action for addition based on the previously branched actions and their current action-value estimates. 

\subsection{Optimal Action Branching}~\label{bayesoptmath}


MCTS with progressive widening operates by incrementally adding actions to the search tree when the branching threshold is met.
For a belief node $b$ with action child nodes $\mathrm{Ch}(b)$, the maximum child node value estimate can be defined as
\begin{equation}
    \hat{Q}(b,a^*) \gets \max_{a \in \mathrm{Ch}(b)} \hat{Q}(b,a)
\end{equation}
where $\hat{Q}(b,a)$ is the action-value estimate maintained in the search tree.
When the search budget is exhausted, the action with the highest estimated action-value is returned. 

The optimal action to add to the tree is the one expected to most improve the maximum value of the child node set. 
In Bayesian optimization, this is commonly referred to as the \emph{Expected Improvement} acquisition function.
We can define the expected improvement of the action-value to be 
\begin{equation}
    \mathrm{EI}(a \mid b) := \mathbb{E}_{Q\sim GP}\Big[|\hat{Q}(b,a) - \hat{Q}(b,a^*)|^+\Big]
\end{equation}
where $|x|^+ = \max(x, 0)$.
Only non-negative differences are used in calculating the expected improvement.
If the expected value of a newly added action is lower than the current maximum, the action will not be selected as the MCTS return value. 

To calculate EI, the expectation is taken with respect to a distribution over the unknown $Q(b,a)$ function. 
To model this distribution, we use a Gaussian Process (GP) conditioned on the values of each previously visited node in the search tree $\{(ba_i,q_{i})\}_{i=1,...,n}$, where $q_i$ is the action-value estimate at belief-action node $i$, $ba_i$ is a vector representing the corresponding belief and action values, and $n$ is the number of action nodes in the tree.

The distribution of any unobserved belief action-value $q^*(b,a)$ is a Gaussian distribution defined by the GP posterior $P(q^*\mid ba^*, \{(ba,q)\}; \theta)$. 
Using this marginal distribution, an analytical expression for the expected improvement at any point can be defined~\cite{jones1998} as 
\begin{equation}
\begin{split}
    \mathrm{EI}(a \mid b) = & |\Delta(ba)|^+ + \sigma(ba)\phi\bigg(\frac{\Delta(ba)}{\sigma(ba)}\bigg) \\ 
    &- |\Delta(ba)|\Phi\bigg(\frac{\Delta(ba)}{\sigma(ba)}\bigg)
\end{split}
\end{equation}
where $\Delta(ba) = \mu(ba) - \hat{Q}(b,a^*)$, $\mu(ba)$ and $\sigma(ba)$ are the GP posterior mean and standard deviation, and $\phi$ and  $\Phi$ are the standard normal probability density and cumulative density functions, respectively. 
The expected improvement can be easily calculated for any candidate action using this closed form and marginal statistics calculated as in~\cref{eq:GP Posterior}.


During action expansion at belief node $b$, the optimal action node 
\begin{equation}\label{eq:argmax}
    \tilde{a} \gets \arg\max_{a \in \mathcal{A}\backslash\mathrm{Ch}(b)} \mathrm{EI}(a \mid b)    
\end{equation}
is added to the tree, and the resulting value estimate $\hat{Q}(b, \tilde{a})$ is added to the GP.
Any appropriate optimizer may be used to solve~\cref{eq:argmax}. 
In the proposed approach, L-BFGS~\cite{liu1989} is used for continuous action spaces and exhaustive search is used for discrete spaces.

\subsection{Algorithm}
We now define Bayesian Optimized Monte Carlo Planning (BOMCP).
The tree search procedures in BOMCP are adapted from the POMCPOW algorithm.
In addition to BOMCP, we also developed Bayesian Optimized Monte Carlo Tree Search (BOMCTS) for fully observable MDPs. 
BOMCTS is not presented in detail in this work, and more information may be found in the Appendix.

\subsubsection{Overview}
The entry point to BOMCP is the \textsc{Plan} procedure, which is shown in~\cref{Plan}.
The algorithm constructs a search tree by repeatedly calling \textsc{Simulate} from the root belief node of the tree for a specified number of times. 
The algorithm then returns the action at the root node with the highest estimated action-value. 

The BOMCP search tree is constructed from alternating layers of belief nodes and action nodes.
This is a departure from POMCP and POMCPOW, which use trees of alternating action and observation nodes, with observation nodes containing state particle collections.
Using beliefs directly allows BOMCP to maintain less biased, though higher variance, sampling of new states during progressive widening steps.
Because BOMCP uses intermediate action-value estimates for the Bayesian optimization process, it was important to minimize the bias on these estimates.
If desired, BOMCP may be configured to use particle collections to represent belief and recover the behavior of POMCPOW.

The majority of the computation occurs in the \textsc{Simulate} procedure.
To begin each query, an initial state is sampled from the root belief and passed to \textsc{Simulate}.
The state and belief are updated during each simulation using the provided generative model, \textsc{Gen}, and update filter, \textsc{UpdateBelief}. 
Both action and belief progressive widening are used to determine expansion at each internal node. 
The UCT algorithm is used to select existing action nodes for exploration when not expanding.

\textsc{Simulate} is called recursively on the updated states and beliefs until a specified maximum depth is reached or the search reaches a leaf node. 
A rollout simulator is used to estimate the leaf node values. 
The leaf-node value estimates are then backed-up the tree along the current history to update the value estimates of the parent nodes. 
\begin{algorithm}[t]
\caption{Plan}\label{Plan} 
\begin{algorithmic}[1]
\Procedure{Plan}{$b$, $B$}
\State $T \gets Node(b)$
\For {$i \in 1 : n$}
    \State $s \sim b$
    \State $\textsc{Simulate}(s, b, d_{max}, T, B)$
\EndFor
\State $B \gets \textsc{UpdateBuffer}(B,T)$
\State \Return $\arg\max_a Q(b,a), \ B$
\EndProcedure
\end{algorithmic}
\end{algorithm}

\begin{algorithm}[h]
\caption{Simulate}\label{simulate} 
\begin{algorithmic}[1]
\Procedure{Simulate}{$s$, $b$, $d$, $T$, $B$}
\If {d = 0}
    \State \Return 0
\EndIf
\If {$|\mathrm{Ch}(b)| \leq k_aN(b)^{\alpha_a}$} 
    \State $a \gets \textsc{BayesOpt}(b, T, B)$
    \State $ \mathrm{Ch}(b) \gets \mathrm{Ch}(b) \cup \{a\}$
    \State $ \mathrm{Ch}(b,a) \gets \emptyset$
\EndIf
\State $a \gets \arg\max_{a\in \mathrm{Ch}(b)} \mathrm{UCB}(b,a)$
\State $s', o, r \gets \textsc{Gen}(s, a)$
\State $New \  Node \gets \mathrm{False}$
\If {$(b,a,b') \in T$}
    \State $M(b,a,b') \gets M(b,a,b') + 1$
\ElsIf {$|\mathrm{Ch}(b, a)| \leq k_bN(b, a)^{\alpha_b}$}
    \State $b' \gets \textsc{UpdateBelief}(b, o)$
    \State $\textsc{InitializeNode}(b,a,b',T)$
    \State $New \ Node \gets \mathrm{True}$
\Else 
    \State $b' \sim P(\mathrm{Ch}(b,a)\mid b, a) \propto M(b', a, b)$ 
    \State $s' \sim b'$
    \State $r \gets R(s, a, s')$
\EndIf
\If {$New \ Node$}
    \State $q \gets r + \gamma \textsc{Rollout}(s', d - 1)$
\Else 
    \State $q \gets r + \gamma \textsc{Simulate}(s', b', d - 1)$
\EndIf
\State $N(b) \gets N(b) + 1$
\State $N(b,a) \gets N(b,a) + 1$
\State $Q(b,a) \gets Q(b,a) + \frac{q - Q(b,a)}{N(b,a)}$
\State \Return q
\EndProcedure
\end{algorithmic}
\end{algorithm}

Bayesian optimization is introduced during action progressive widening.
When a new action is added to the search tree, instead of selecting the action randomly from the accessible action space, BOMCP uses the procedure defined in~\cref{BayesOpt}.
This routine constructs a Gaussian Process using all previously visited belief-action node pairs $(b, a)$ and corresponding $Q(b,a)$ estimates currently in the search tree $T$ and buffer $B$. 
It then returns the action that maximizes the expected improvement.
\begin{algorithm}[b]
\caption{Bayesian Optimization}\label{BayesOpt} 
\begin{algorithmic}[1]
\Procedure{BayesOpt}{$b$, $T$, $B$}

\State $X \gets ()$
\State $Y \gets ()$
\For {$(b,a) \in T \cup B$}
    \State $\text{append} \ \textsc{Vectorize}(b,a) \ \text{to} \ X$
    \State $\text{append} \ Q(b,a) \ \text{to} \ Y$
\EndFor
\State $gp \gets \textsc{GaussianProcess}(X, Y)$
\State $X^* \gets \textsc{Actions}(b) \backslash X$
\State $\mu^*, \sigma^* \gets \textsc{Posterior}(gp, X^*)$
\State \Return $\arg\max_a \mathrm{EI}(\mu^*, \sigma^*)$
\EndProcedure
\end{algorithmic}
\end{algorithm}

\subsubsection{Gaussian Process}
The Gaussian Process is defined by the kernel, prior mean, and additive noise. 
In BOMCP, these are treated as hyper-parameters and must be specified. 
By default, BOMCP uses a squared-exponential kernel of the form
\begin{equation}
    k_{SE}(x_1, x_2) = \sigma^2 \mathrm{exp}\bigg(-\frac{(x_1 - x_2)^2}{2\ell^2}\bigg)
\end{equation}
where $\sigma^2$ is the marginal variance, and $\ell$ is the characteristic length. 
Larger characteristic lengths imply stronger correlation between belief-action values. 

Using a constant prior mean $\mu_0$ is common in Bayesian optimization and is the default setting for BOMCP.
The $\mu_0$ value can be used to tune the exploration behavior.
Setting $\mu_0$ to a lower bound on the action-value leads to conservative branching, with actions tending to be selected closer to the previously observed optimum, while setting $\mu_0$ to an upper bound tends to encourage more optimistic exploration.

In order to use the Gaussian Process, the inputs must be represented as vectors. 
A function \textsc{Vectorize} must be defined to transform belief-action pairs $(b,a)$ to vector representations, $ba$. 
For high-dimensional belief spaces, it may be beneficial for \textsc{Vectorize} to represent the belief as a reduced dimensional collection of sufficient statistics. 

In general, Gaussian Process inference is not tractable for large data-sets, as it is $O(n^3)$ in the number of prior observations~\cite{rasmussen2005}.
This cubic dependence is caused by the need to invert the $n$-by-$n$ covariance matrix.
Several approximate methods have been proposed to reduce the size of this inversion~\cite{silverman1985}.

BOMCP uses a $k$-nearest neighbor approach inspired by Sequential Gaussian Simulation~\cite{gomez1993}, which only considers the effects of the nearest observed points for each posterior calculation. 
This replaces the $n$-by-$n$ inversion with a series of $k$-by-$k$ inversions, where $k \ll n$.
The resulting approximation is similar to existing nearest-neighbor Gaussian Process regression techniques~\cite{datta2016}.
Additional details on this process may be found in the Appendix.

\subsubsection{Experience Buffer}
The actions added at the root of the tree are the most important to the performance of the search, since one of these actions will be be returned.
We would like to ensure good actions are added to the tree at the root. 
The Bayesian optimization procedure requires value estimates from existing nodes to be effective.
Unfortunately, early in the search process, when many of the root actions are added, there are very few nodes.
To offset this, we introduced an experience buffer, $B$, containing belief-action nodes from prior planning steps.
The observations from this buffer are used with the current tree nodes to build the Gaussian Process. 
The buffer is stochastically re-filled at the end of \textsc{Plan}.


\section{Experiments}
We implemented BOMCP and BOMCTS in Julia building upon the POMDPs.jl package~\cite{egorov2017}.
To evaluate the effectiveness of BOMCP, we conducted a series of experiments on three distinct POMDPs. 
We evaluated the performance of BOMCP against the performance of POMCPOW and expert policies for each problem. 

For each experiment, we recorded the task score as well as the wall clock run time per-search to measure the computation cost. 
We ran each experiment with varying numbers of queries-per-search.
For all tests, the same values were used for hyper-parameters shared by BOMCP and POMCPOW such as $K_{action}$ and $\alpha_{action}$.
Source code for BOMCP and the tasks is provided in the supplemental materials.

\subsection{Partially Observable Lunar Lander}
\begin{figure}
     \centering
     \includegraphics[width=0.9\columnwidth]{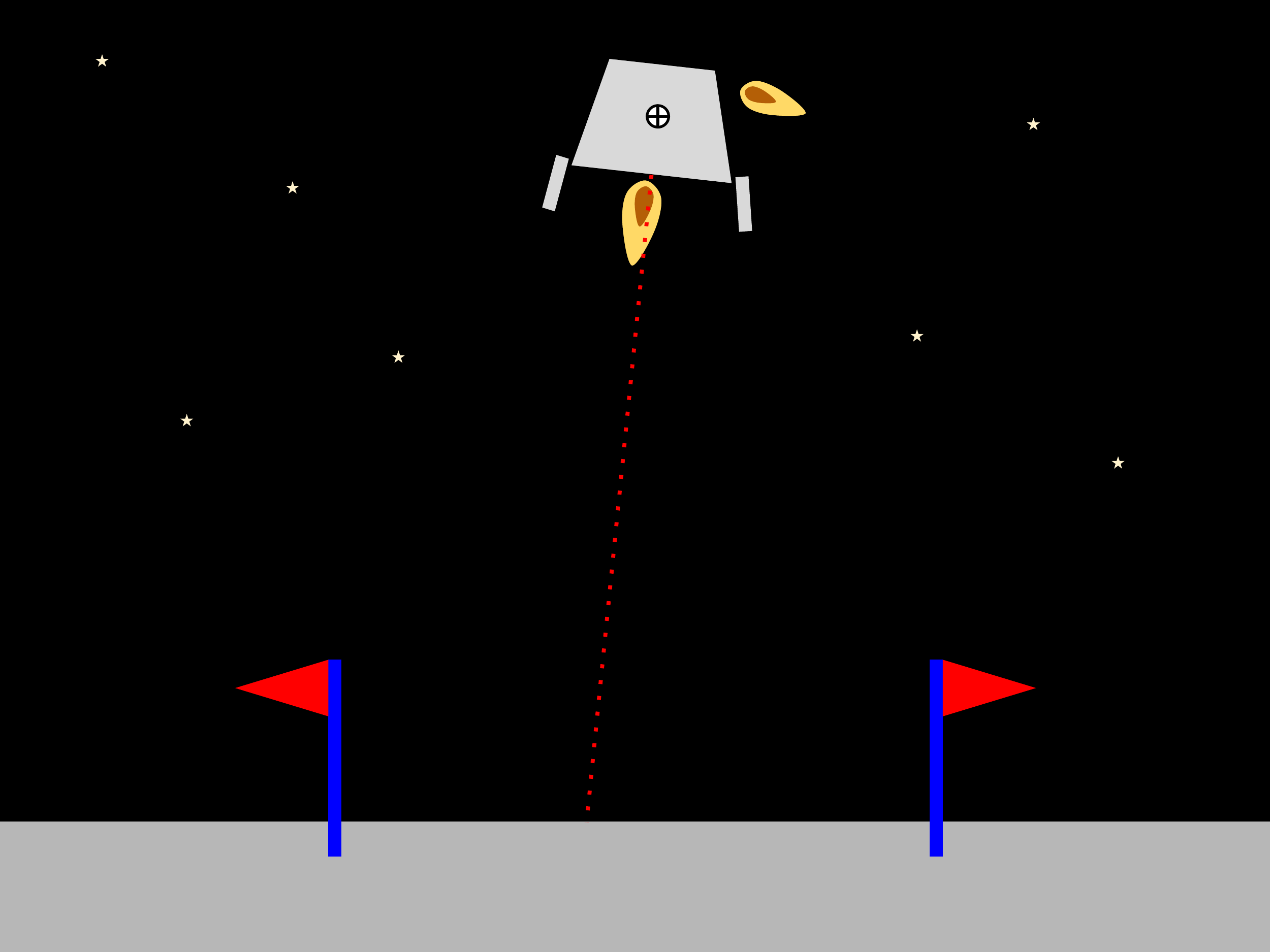}
     \caption{Partially Observable Lunar Lander. 
    Control is applied through vertical and lateral thrust, depicted as the flames. Noisy observations of the altitude give the distance from the vehicle's center to the floor along the vertical axis of the lander, as depicted by the dashed red line.}
     \label{fig:lander}
 \end{figure}

The first problem studied was a partially-observable variant of the popular lunar-lander problem.
The objective of the task is to guide a vehicle to land in a target area with low impact force.
The environment is shown in~\cref{fig:lander}.

The vehicle state is represented by a six dimensional tuple $(x,y,\theta, \dot{x}, \dot{y}, \omega)$, where $x$ and $y$ are the horizontal and vertical positions, $\theta$ is the orientation angle, $\dot{x}$ and $\dot{y}$ are the horizontal and vertical speeds, and $\omega$ is the angular rate.

The vehicle makes noisy observations of its angular rate, horizontal speed, and above-ground level (AGL), which is represented as the dashed red line in~\cref{fig:lander}. 
The action space is a three-dimensional continuous space defined by the tuple $(T, F_x, \delta)$.
$T$ is the main thrust which acts along the vehicle's vertical axis through its center of mass, and is in the range $[0,15]$.
$F_x$ is the corrective thrust, which acts along a horizontal axis offset from the center of mass by a distance $\delta$.
$F_x$ is in the range $[-5, 5]$ and $\delta \in [-1, 1]$. 

The initial vehicle state is sampled from a multivariate Gaussian with mean $\mu=(x=0, y=50, \theta=0, \dot{x}=0, \dot{y}=-10, \omega=0)$. 
The reward function is defined as 
\begin{equation}
r(s,a,s') =  
\begin{cases}
    -1000,& \text{if } x \geq 15 \vee \theta \geq 0.5\\
    100 - x - v_y^2,& \text{if } y \leq 1\\
    -1,              & \text{otherwise}
\end{cases}
\end{equation}
The first term in the reward function provides a penalty for the vehicle entering an unrecoverable state.
The second term provides a positive reward for landing the vehicle, minus a penalty for drift away from center and for impact velocity. 
The final term is a constant penalty for fuel consumption.

We tested BOMCP and POMCPOW with varying numbers of queries per tree search. 
An Extended Kalman Filter was used to maintain a multi-variate Gaussian belief over vehicle state.
An expert policy was also tested. 
This expert policy was also used as the rollout policy for leafnode evaluations for both BOMCP and POMCPOW.
The results are summarized in~\cref{tab:lander}. 
\begin{table}[h]
    \caption{Lunar Lander Results. The mean and one standard error bound are given for the episode score and per-search wall clock runtime.}~\label{tab:lander} 
    \centering
    \begin{tabular}{@{}lrrr@{}}
        \toprule
        Algorithm & Queries & Score & Time (seconds) \\
        \midrule
        \multirow{4}{*}{POMCPOW} & 10 & $-92 \pm 9$ & $0.003 \pm 0.001$\\
        & 50 & $-5 \pm 6$ & $0.009 \pm 0.002$ \\
        & 100 & $13 \pm 5$ & $0.013 \pm 0.007$ \\
        & 500 & $31 \pm 4$ & $0.051 \pm 0.013$ \\
        & 1000 & $32 \pm 3$ & $0.130 \pm 0.012$ \\ 
        \midrule
        \multirow{4}{*}{BOMCP} & 10 & $-50 \pm 1$ & $0.028 \pm 0.001$\\
        & 50 & $36 \pm 1$ & $0.070 \pm 0.011$ \\
        & 100 & $57 \pm 2$ & $0.141 \pm 0.010$ \\
        & 500 & $61 \pm 2$ & $0.727 \pm 0.017$ \\
        \midrule
        Expert & -- & $-320 \pm 28$ & -- \\
        \bottomrule
    \end{tabular}
\end{table}

For each number of queries-per-search tested, BOMCP outperforms POMCPOW by a statistically significant margin.
Despite the low-dimensional action-space, optimal action selection still significantly improves performance in BOMCP.

BOMCP takes significantly longer per-search than POMCPOW. 
Profiling BOMCP \textsc{Plan} revealed that approximately $30\%$ of the time was spent in action selection. 
In POMCPOW, a negligible amount of time was spent in action selection. 

We ran an additional set of tests on POMCPOW for 1000 queries-per-search.
At this number of queries, POMCPOW took as much time per-search as BOMCP with 100 queries.
BOMCP at 100 queries still outperforms POMCPOW at 1000, with both algorithms taking approximately $0.13$ seconds per-search.
This suggests BOMCP outperforms POMCPOW by selecting better actions, rather than by just using additional computation.

\subsection{Wind Farm Planning} 
The second task was large-scale wind farm planning.
In this task, the agent sequentially selects locations to install sensor towers in a large three-dimensional wind field. 
The objective is to generate accurate maps of high wind areas which will later be used to plan turbine layouts. 

\begin{figure}[h!]
\centering
\includegraphics[width=0.9\columnwidth, trim={0.8cm 0.8cm 0.7cm 0.8cm},clip]{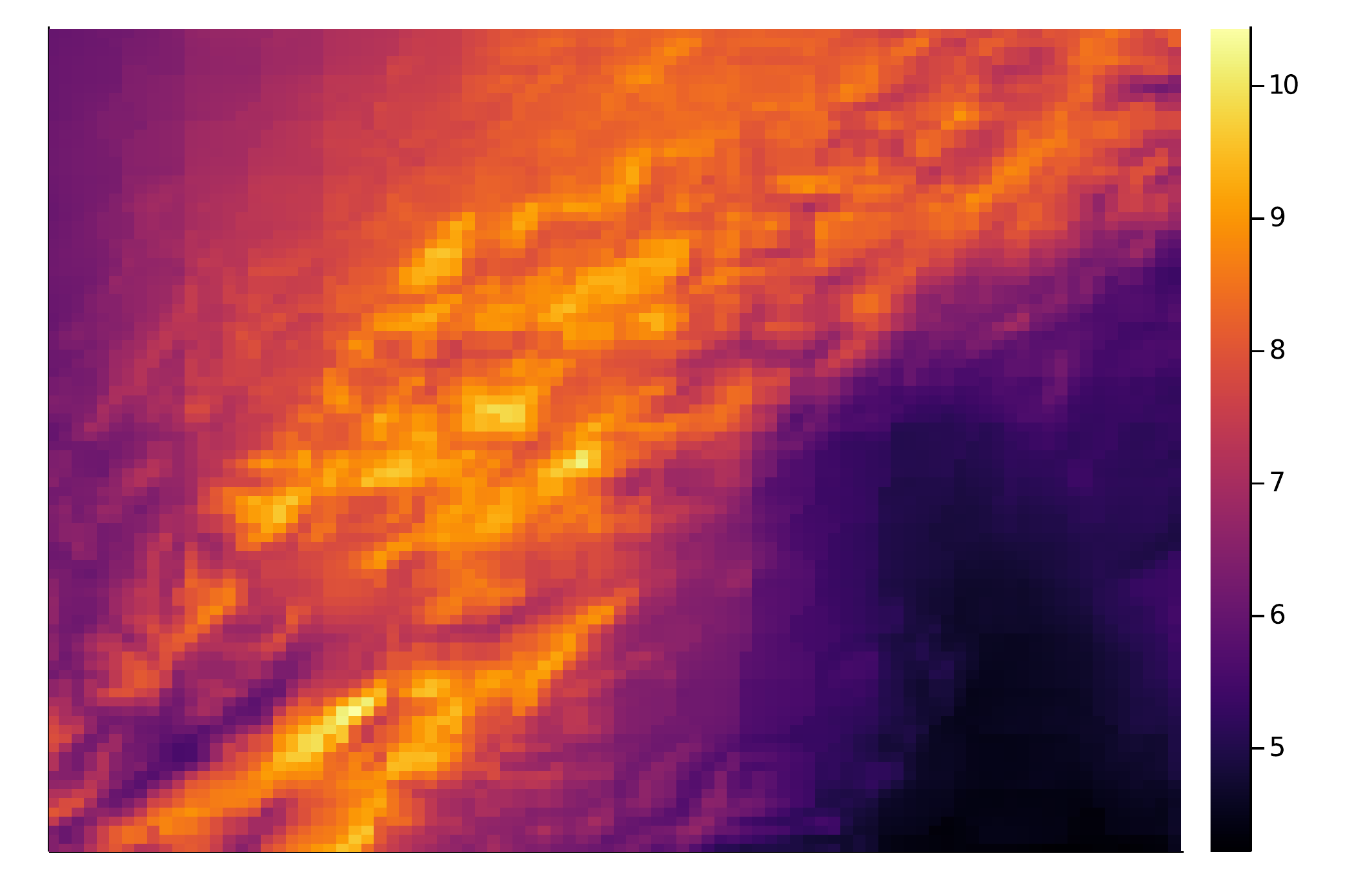}
\caption{Wind Map. Figure shows wind map for Altamont Pass, CA at $100$m altitude. The colors represent the average annual wind speed in m/s.}
\label{fig:altamont}
\end{figure}

The environment state is represented by a three-dimensional wind map of average annual wind-speed at discrete locations in the wind field. 
We used data from the Global Wind Atlas at the Altamont Pass wind farm, which covers an area of approximately 
\SI{392}{\square\km}. 
The wind-field at a single altitude is shown in~\cref{fig:altamont}.

We selected a sub-region of the field for our experiments.
The map of this region was a $20 \times 20 \times 3$ array, covering a \SI{4440}{\metre} $\times$ \SI{4440}{\metre} area at altitudes \SI{50}{\metre}, \SI{100}{\metre}, and \SI{150}{\metre}.

In addition to the wind map, the state also contains the location of all previously placed sensor towers.

At each time step, the agent may choose to place a sensor tower at any unoccupied grid location on the map, at any of the three altitudes. 
The action space, therefore, contains at most 1200 distinct actions.
The agent receives a noiseless observation of the wind value at each sensor tower location, but nowhere else.
A sensor tower observes wind values all altitudes at and below its height. 
It is assumed that some initial knowledge of the wind field is available, represented as a set of sparse prior observations in a Gaussian Process.

Reward is generated by first passing the current belief to a turbine-layout optimizer. 
The optimizer produces a risk-sensitive turbine arrangement by greedily selecting locations with the maximum $1-\sigma$ lower confidence bound.
The resulting layout is then used to estimate total annual power production based on the true wind field state. 
An additional cost is incurred for each tower, 
linearly proportional to its height.
In this way, the reward encourages sensor arrangements that reduce variance in areas of high wind. 

As with lunar lander, we tested BOMCP and POMCPOW with varying numbers of queries per tree search for 200 trials each. 
Mean episode returns with one standard error bounds are shown in~\cref{tab:wind_results}.


\begin{table}[h]
    \caption{Wind Farm Planning Results. The mean and one standard error bound are given for the episode score and per-search wall clock runtime.}~\label{tab:wind_results} 
    \centering
    \begin{tabular}{@{}lrrr@{}}
        \toprule
        Algorithm & Queries & Score & Time (seconds) \\
        \midrule
        \multirow{5}{*}{POMCPOW}  & 10 & $15708 \pm 229$ & $2.25 \pm 0.07$ \\
        & 25 & $16234 \pm 217$ & $4.80 \pm 0.07$ \\
        & 50 & $16374\pm 212$ & $6.27 \pm 0.08$ \\
        & 100 & $16018 \pm 262$ & $11.98 \pm 0.07$ \\
        & 200 & $15787 \pm 233$ & $20.67 \pm 0.09$ \\
        \midrule
        \multirow{5}{*}{BOMCP} & 10 & $ 18095 \pm 183$ & $2.55 \pm 0.08$ \\
        & 25 & $ 18154 \pm 158$ & $5.21 \pm 0.07$ \\
        & 50 & $ 18015 \pm 163$ & $6.71 \pm 0.06$ \\
        & 100 & $ 18225 \pm 119$ & $13.39 \pm 0.07$ \\
        & 200 & $ 18113 \pm 157$ & $25.14 \pm 0.08$ \\
        \midrule
        Expert & -- & $ 8130 \pm 51$ & -- \\
        \bottomrule
    \end{tabular}
\end{table}

For this large, discrete action-space problem, BOMCP significantly outperformed POMCPOW. 
Both BOMCP and POMCPOW seem to reach maximum performance at approximately 50 queries. 
POMCPOW's performance never exceeds the performance of BOMCP with even just 1 query. 
Somewhat surprisingly, with a single query each, BOMCP still out performs POMCPOW.
This is likely due to the BOMCP experience buffer, which allows BOMCP to choose intelligent actions while POMCPOW samples randomly. 

\subsection{Cyber Security}
The final task was a network cyber security task in which the agent scans nodes on a network in order to detect and eliminate a spreading malware infection. 
This task was selected because a distance metric between Gaussian Process observations was not as clearly apparent.
The problem tests the robustness of BOMCP to more difficult to represent spaces. 

The network is composed of a set of Local Area Networks (LANs), where each LAN is a set of fully connected host nodes and a single server node.
Edges between server nodes are generated randomly, though the graph is constrained to be complete. 
The network also contains one special ``vault" server node, which is the malware target.

The state is represented by the infection states of the individual nodes as well as the edge topology. 
There are $4$ LANs and $10$ host nodes per LAN.
The infection spreads according to a known stochastic adversary policy.

Each step, the agent may scan every node on a single LAN or scan an individual node. 
When scanning a LAN, the agent detects malware on all LAN nodes with probability $0.3$.
When scanning single nodes, the agent detects malware with probability $0.5$ and cleans it with probability $0.8$.

The episode terminates if the vault node is infected or 250 timesteps have passed. 
The agent receives no reward on non-terminal steps. 
For terminal steps, the reward function is
\begin{equation}
r(s,a,s') =  
\begin{cases}
    -100 - 0.5|S_i| - 0.1|H_i|,& \text{if vault infected} \\
    -0.5|S_i| - 0.1|H_i|,              & \text{otherwise}
\end{cases}
\end{equation}
where $S_i$ is the set of infected server nodes and $H_i$ is the set of infected host nodes.

A Dynamic Bayes Network was used to represent the belief model, with a discrete Bayes filter used for updates. 
The results of experiments with BOMCP and POMCPOW are summarized in~\cref{tab:cyber}. 

As with the previous experiments, BOMCP outperforms POMCPOW for a given number of queries, though at significantly higher cost per query.
Despite the lack of a physical distance to use for the Gaussian Process kernel, BOMCP still offered significant improvements over POMCPOW.
\begin{table}[h]
    \caption{Cyber Security Results. The mean and one standard error bound are given for the episode score and per-search wall clock runtime.}~\label{tab:cyber} 
    \centering
    \begin{tabular}{@{}lrrr@{}}
        \toprule
        Algorithm & Queries & Score & Time (seconds) \\
        \midrule
        \multirow{3}{*}{POMCPOW} & 50 & $-62 \pm 4$ & $0.13 \pm 0.07$\\
        & 100 & $-49 \pm 5$ & $0.23 \pm 0.07$ \\
        & 500 & $-31 \pm 4$ & $1.01 \pm 0.22$ \\
        \midrule
        \multirow{3}{*}{BOMCP} & 50 & $-27 \pm 4$ & $0.71 \pm 0.14$\\
        & 100 & $-23 \pm 4$ & $1.38 \pm 0.27$ \\
        & 500 & $-23 \pm 4$ & $6.12 \pm 0.49$ \\
        \midrule
        Expert & -- & $-55 \pm 4$ & -- \\
        \bottomrule
    \end{tabular}
\end{table}
\section{Conclusions}
In this work, we framed MCTS action selection as a Bayesian optimization problem. 
We solved this optimization problem in a new POMDP planning algorithm called BOMCP. 
Experiments showed that BOMCP significantly outperforms POMCPOW on a variety of large scale POMDPs, though at much higher computational cost.
These results suggest that the current implementation of BOMCP may be best suited for problems with computationally expensive rollout simulators. 

This work demonstrated significant improvements to MCTS through optimized action branching.
The work also showed that direct application of Bayesian optimization can result in significant increase in computational cost, even with Gaussian Process approximations. 
Future work will explore additional methods to reduce computational cost, including use of different distributional models. 

Future work will also investigate different formulations of the Bayesian Optimization problem. 
Alternative acquisition functions to Expected Improvement will be considered. 
In some instances, tuning the Gaussian Process hyper-parameters required significant trial and error.
Methods to dynamically tune these parameters will also be developed. 

Using the Gaussian Process distribution over $Q(b,a)$ to generalize a search tree to unobserved trajectories will also be explored to allow it to be used in an offline setting. 

\bibliography{bibliography.bib}
\bibstyle{aaai21}
\end{document}